\documentclass[10pt,journal,compsoc]{IEEEtran}
\usepackage{graphicx} 
\usepackage{multirow}
\usepackage{algorithm}
\usepackage{algpseudocode}
\usepackage{caption}
\usepackage{subcaption}
\usepackage{url}
\usepackage{amsmath,amssymb,amsfonts}
\usepackage{textcomp}

\ifCLASSOPTIONcompsoc
  \usepackage[nocompress]{cite}
\else
  \usepackage{cite}
\fi
\ifCLASSINFOpdf
\else
\fi
\ifCLASSOPTIONcompsoc
    \usepackage{color}
    \usepackage{comment}
    \newif\ifshowcomments
    \showcommentstrue 

\fi

\usepackage{fancyhdr}
\fancypagestyle{firstpage}{
\fancyhf{}
\fancyfoot[L]{This preprint has been further published in IEEE Access \copyright IEEE. Print ISSN: 2169-3536. Online ISSN: 2169-3536. Digital Object Identifier: 10.1109/ACCESS.2025.3548957}

}

\begin{document}
%
\title{SynMorph: Generating Synthetic Face Morphing Dataset with Mated Samples}
%
%
%
%


\author{Haoyu Zhang$^\dag$, Raghavendra Ramachandra$^\dag$ \\ Kiran Raja$^\dag$,  Christoph Busch$^\dag$$^\ddagger$ \\
$^\dag$ Norwegian University of Science and Technology (NTNU), Norway\\  
	$^\ddagger$Darmstadt University of Applied Sciences (HDA), Germany\\
	\{\tt\small haoyu.zhang; raghavendra.ramachandra;kiran.raja;christoph.busch\} @ntnu.no\\
	\{\tt\small christoph.busch\}@h-da.de\\
}

\IEEEtitleabstractindextext{%
\begin{abstract}
Face morphing attack detection (MAD) algorithms have become essential to overcome the vulnerability of face recognition systems. To solve the lack of large-scale and public-available datasets due to privacy concerns and restrictions, in this work we propose a new method to generate a synthetic face morphing dataset with 2450 identities and more than 100k morphs. The proposed synthetic face morphing dataset is unique for its high-quality samples, different types of morphing algorithms, and the generalization for both single and differential morphing attack detection algorithms. For experiments, we apply face image quality assessment and vulnerability analysis to evaluate the proposed synthetic face morphing dataset from the perspective of biometric sample quality and morphing attack potential on face recognition systems. The results are benchmarked with an existing SOTA synthetic dataset and a representative non-synthetic and indicate improvement compared with the SOTA. Additionally, we design different protocols and study the applicability of using the proposed synthetic dataset on training morphing attack detection algorithms.

\end{abstract}

}

\maketitle
\thispagestyle{firstpage}
\IEEEdisplaynontitleabstractindextext

%
\IEEEpeerreviewmaketitle

\IEEEraisesectionheading{\section{Introduction}\label{sec:introduction}}

%
%
%
%
 \IEEEPARstart{F}{ace} recognition systems(FRS) have been widely deployed in different secure application scenarios, such as automatic border control \cite{venkatesh2021face}. Nonetheless, with the improvement of FRS in generalization and the development of image manipulation techniques, it is also shown that FRS is vulnerable to various types of attacks \cite{ngan2024face} \cite{raja2020morphing}. Hence, it is essential to develop corresponding attack detection algorithms to protect the FRS from potential attacks. Face morphing attack detection (MAD)  is the technology detecting attacks that combine the face images from two or more individuals into a single morphed image. Based on the attack scenario and the types of input, MAD can be classified into single image-based morphing attack detection (S-MAD) and differential image-based morphing attack detection (D-MAD). S-MAD aims to detect the face morphing attack based on a single image presented to the algorithm. The common application scenario is validating photos submitted during the application for a visa or passport and validating the existing database without morphed images. The D-MAD case simulates the scenario of automatic border control, where a suspicious image in the passport is validated, given the supplementary information from trustworthy probes captured by the gate cameras.

Various MAD approaches have been designed by researchers \cite{venkatesh2021face}. Additionally, based on their approach, they can be roughly classified into explicit methods using engineered features such as hand-crafted texture descriptors and implicit methods with advanced deep learning techniques that can achieve better generalizability. In both cases, most of the algorithms are data-driven and while the former offers some explainability, the latter needs a larger size of the training data to avoid overfitting.  Hence, it is essential to have large-scale and high-quality training datasets to develop generalized and robust MAD algorithms and testing datasets to evaluate and benchmark existing algorithms from different developers. However, due to privacy regulations, face samples are considered sensitive data, which makes it challenging to collect the dataset on a large scale and difficult to share between researchers in different institutes.

Several works have been done to address this challenge. The most common solution is benchmarking different algorithms with in-house protocol and database. However, as the dataset is not publicly available, it lacks transparency. Meanwhile, this will make the results from different research work challenging to compare and, hence, less reproducible. Another existing solution is benchmarking MAD algorithms in public evaluation platforms such as NIST FATE MORPH \cite{ngan2024face}, and Bologna Online Evaluation Platform (BOEP) \cite{raja2020morphing} \footnote{\url{https://biolab.csr.unibo.it/fvcongoing/UI/Form/BOEP.aspx}}. In this way, trained algorithms are submitted to the evaluation platforms and benchmarked with other submitted algorithms. However, submitting to these platforms requires following a specific application programming interface, which is not convenient for all of the approaches and their implementations. Further, the training phase of the submitted algorithms is not transparent between different algorithms as the developers use their own training data.

A convenient approach is to use a transparent, sharable synthetic dataset that can scale into a large number of samples. Several works have been conducted to design algorithms for the generation of synthetic 2D face data \cite{colbois2021use} \cite{melzi2023gandiffface} and evaluate the applicability of synthetic data in training and testing face recognition systems. However, the task of generating synthetic data for MAD poses two specific criteria: 1) Realism - face morphing attack detection is often based on detailed traces created by the morphing process distributed on the face region. Compared to tasks understanding visual content, this may increase the gap between synthetic and non-synthetic data. 2) Representativeness - as an application-oriented task, face morphing attack detection has a common application scenario (e.g., the suspicious image should be similar to passport quality and the probe image should not be completely in-the-wild), hence, it requires specialised algorithms for generating synthetic datasets for MAD tasks.

\begin{figure*}[htbp]
    \centering
    \includegraphics[scale=0.7]{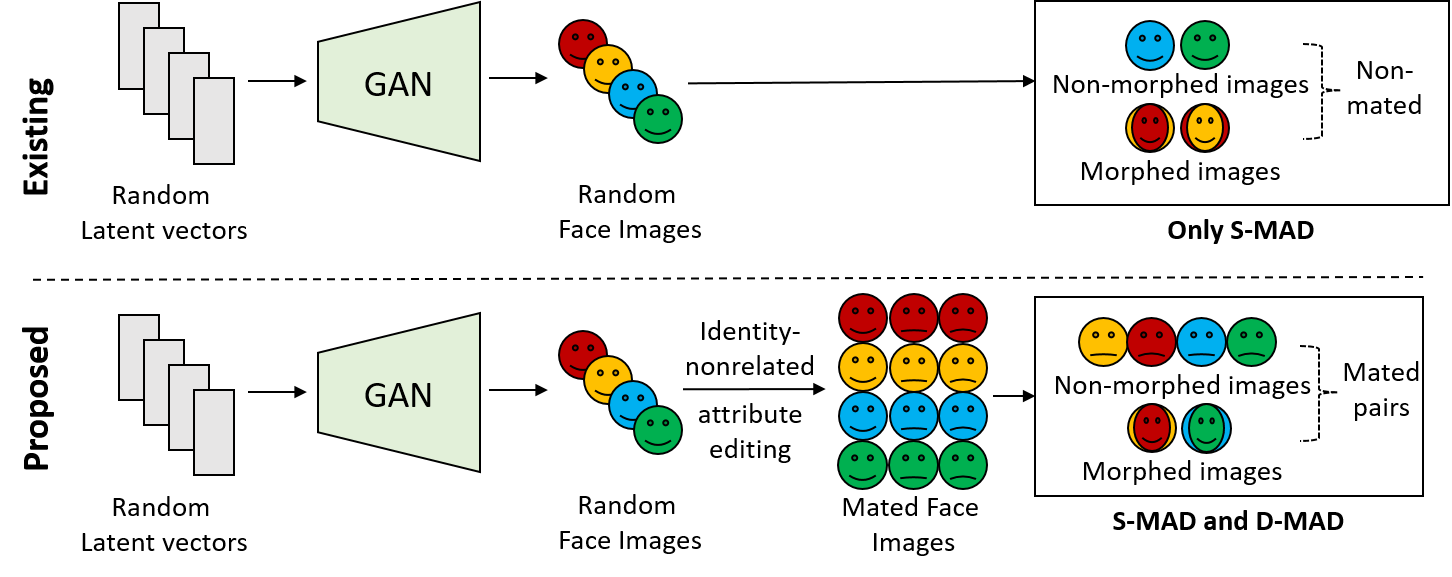}
    \caption{Overall summary and comparison of existing approaches and proposed SynMorph approach for generating synthetic morph dataset.}
    \label{fig:motivation}
\end{figure*}

Existing approaches are using randomly sampled latent vectors to create face images with assumed different identities. Damer et al. \cite{damer2022privacy} proposed a method to generate a synthetic morphing dataset. In this work, random face samples are generated by the StyleGAN \cite{karras2019style} model in the resolution of 256 $\times$ 256. To exclude images with low biometric sample quality, an end-to-end face image quality assessment (FIQA) algorithm to predict the recognition performances of generated synthetic data was employed. Then, a FRS is applied to select similar pairs of images contributing to the morphs. Finally, morphed images are generated using landmark-based algorithms. The dataset is named as Synthetic Morphing Attack Detection Development dataset (SMDD) dataset. Later, Tapia et al. \cite{tapia2023impact} extended the dataset with more morphing algorithms and conducted cross-dataset testing. However, as the dataset only contains randomly generated images as random identities, there's no mated sample included. Hence the dataset only supports training and testing of S-MAD algorithms. We note the following limitations from existing works: 1) the face image quality is restricted for the implicit FIQA filtering and small image resolution  2) only one landmark-based morphing algorithm without any post-processing procedure is applied to generate the morphs in synthetic dataset 3) without mated samples included, the dataset only supports the development of S-MAD algorithms. 

Motivated by this, in this paper we present a novel approach to generate a high-quality synthetic morphing dataset that supports both S-MAD and D-MAD applications as illustrated in Figure \ref{fig:motivation}. To improve the face image quality of the generated synthetic data, StyleGAN 2 model \cite{karras2020analyzing} pre-trained on the FFHQ dataset with 1024 $\times$ 1024 image resolution is applied. Meanwhile, explicit face quality measures (neutral pose, no occlusion) are applied to filter out samples of interest. Furthermore, latent editing techniques are used to neutralize expression and illumination conditions instead of randomly generated images. For the morphs, we use one GAN-based morphing algorithm \cite{zhang2021mipgan} and one landmark-based morphing algorithm with post-processing \cite{UBO_Morphing_Tool} to generate the morphs. This enables the study of cross-morphing-attack, and the SOTA algorithm can generate challenging morphed images and hence improve the robustness of the MAD algorithm trained on this dataset. To create a face morphing dataset that supports both S-MAD and D-MAD cases, we propose to generate the mated samples by editing face attributes in different configurations. The different editing configurations will result in mated samples for S-MAD and D-MAD cases, respectively. In this way, we use the proposed method to generate a synthetic face morphing dataset with an image resolution of 1024 $\times$ 1024, over 100k samples for each morph subset and non-morph subset. Upon acceptance of this paper, the dataset will be published. Further, the dataset is evaluated from the perspective of FIQA, vulnerability analysis, and training of MAD algorithms. The following summarizes the contribution of this work:
\begin{itemize}
    \item A new approach is proposed for generating a high-quality synthetic morphing dataset that supports both S-MAD and D-MAD development.
    \item A high-quality synthetic morphing dataset generated by the proposed method is presented. In total 2450 subjects and in total 500k samples are included in this dataset. Dataset is available for the research purpose. \footnote{\url{https://share.nbl.nislab.no/HaoyuZhang/SynMorph_public}}
    \item Quantitative evaluation results of the generated dataset are reported from the perspective of face image quality and the standardized measurement of morphing attack potential. 
    \item A Study on the applicability of using synthetic datasets for developing S-MAD and D-MAD algorithms with various protocols is conducted.
\end{itemize}

The rest of the paper is organized as follows: Section \ref{sec:related} presents related works on public-available MAD datasets, Section \ref{sec:proposed} presents the proposed method of generating synthetic morph dataset, Section \ref{sec:experiment} first presents the detailed information of our generated synthetic face morphing dataset and the selected baseline synthetic/non-synthetic datasets to be benchmarked. The section also discuss presents our experiments to evaluate the proposed method and the generated synthetic morphing dataset from different perspectives. Section \ref{sec:discussion} discusses the results and overall applicability of synthetic samples. Section \ref{sec:conclusion} draws the overall conclusions.

\section{Related Works}
\label{sec:related}
Existing non-synthetic face morphing datasets are usually constructed based on FRLL \cite{debruine_jones_2017}, FRGC v2.0 \cite{FRGC_DB}, Color FERET \cite{phillips1998feret}, Utrecht ECVP \cite{utrecht_ecvp}, Casia-webface \cite{yi2014learning} or other in-house datasets. The common challenge is that the morph datasets generated by most of these datasets are not publicly sharable for benchmarking MAD algorithms. FRLL-Morph \cite{Sarkar2020} is the existing public-available face morphing dataset, while only 102 subjects and one mated sample for each subject are included in the dataset. Hence the number of morphed samples and non-morphed samples are heavily unbalanced. This indicates another challenge of face morphing dataset: it is challenging to construct a face morphing dataset with both high quality and sufficient size of data for training generalized MAD algorithms.

As several works have been studying the applicability of using synthetic data for training and evaluating face recognition task \cite{boutros2023synthetic}, Damer et al. \cite{damer2022privacy} proposed a method to generate a synthetic face morphing dataset. In their SMDD dataset, non-morphed images are generated using the StyleGAN \cite{karras2019style} model in 256 x 256 image resolution. FaceQnet v1 \cite{hernandez2019faceqnet} algorithm is used to assess the biometric sample quality of generated synthetic images and exclude the ones with low-quality scores. Then, in total 50k non-morphed synthetic images are generated and split into 25k non-morphed images and 25k are used for generating 25k landmark-based morphs. Finally, the generated morphs are filtered again by FIQA, and resulting in a dataset with 25k non-morphed images and 15k morphed images.

\section{Proposed Method}
\label{sec:proposed}
\begin{figure*}
    \centering
    \includegraphics[scale=0.60]{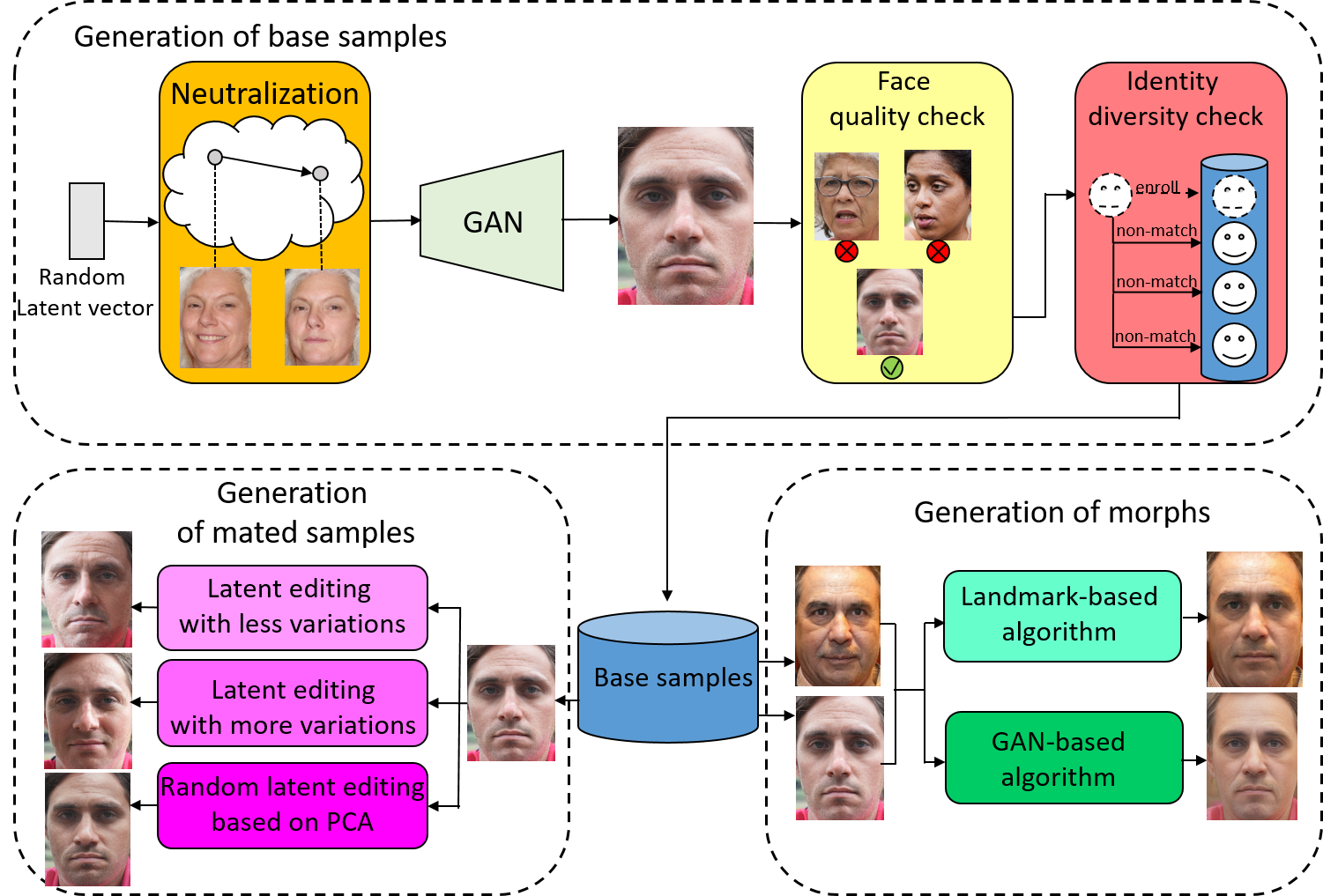}
    \caption{Overview of the generation of SynMorph dataset.}
    \label{fig:overview}
\end{figure*}

In this section, we introduce the proposed method for generating a synthetic face morphing dataset. As shown in Figure \ref{fig:overview}, the method can be divided into three parts: generation of base samples, generation of mated samples, and generation of morphs. First of all, the base samples here denote the images representing different identities in the dataset and will be used to generate mated samples with the same identity. More specifically, base samples are controlled with face image quality and intra-identity diversity so that each base sample aims to represent a high-quality face image of a unique identity among the dataset. Then, corresponding mated samples are generated by applying different attribute editing techniques to the original base sample. Finally, for each base sample, paired base samples for morphing are selected based on similarity and two morphing algorithms are applied to generate the morphed samples.

\subsection{Generation of Base Samples}
For the generation of synthetic images, we use the StyleGAN2 \cite{karras2020analyzing} model pre-trained on the FFHQ dataset \cite{karras2019style}. A pre-trained StyleGAN2 generator maps between a known latent space and the pixel space. Hence, by randomly sampling latent vectors $z \sim \mathcal{N}(0,1)$ in the known distribution, random different faces can be generated. Following the architecture of the StyleGAN2 model, the random sampled latent vector from $\mathcal{Z}$-space will be further mapped by the pre-trained mapping network $f$ to $\mathcal{W}$-space as $w=f(z)$. To simulate the construction of a face morphing database, our target is to generate images that have acceptable face quality for the enrolment process of the passports and diverse identity information between different random images. 
To ensure the face image quality of the accepted samples, we first apply a latent editing technique to neutralize the random sample and then use an explicit face quality filtering pipeline to filter out non-interesting images. The neutralization process was proposed by Colbois et al. \cite{colbois2021use}, where the author proposed to use semantically controlled non-synthetic data to compute the corresponding linear shifting that is required in the latent space to achieve the neutralization of a synthetic image. By fitting SVM classifier for binary attribute classification, the unit normal vector $\hat{n}$ of the SVM's hyperplane is computed as the shifting direction in latent space, and the mean distance $d$ of sample points in each class is calculated as the scale for editing to each corresponding class. More specifically, the sample is edited to have a frontal pose angle, neutralized expression, and neutralized illumination conditions. The binary classes of the pose are based on left or right poses and the binary classes of the illumination are based on light flashed from left or right. Hence the neutralization is sequentially projecting the $\mathcal{W}$-latent vector to corresponding decision boundary as $w' = w - (w^\top \hat{n}_P) \cdot \hat{n}_P$ and $w'' = w' - (w'^\top \hat{n}_I) \cdot \hat{n}_I$, where $\hat{n}_P$ and $\hat{n}_I$ are unit normal vectors of the decision boundary for classifying pose and illumination respectively. The binary classification of expression is between neutral and smiling expressions, hence the $\mathcal{W}$-latent needs to be first projected to the decision boundary using the unit normal vector $\hat{n}_{NS}$ and then shifted with the pre-computed mean distance $d_{NS}^N$ towards the neutral class as $w''' = w'' - (w''^\top \hat{n}_{NS} + d^{N}_{NS}) \cdot \hat{n}_{NS}$. In the further filtering pipeline, we apply img2pose \cite{albiero2021img2pose} to determine the yaw and pitch angle and only accept within the range of $[-5,5]$ degree for both yaw and pitch angles, and then use Dlib \cite{dlib} landmark detection and canny edge detection operator on the bridge of the nose to filter out face images with closed eyes or covered by glasses.
To enrich the diversity of identity information sampled in our database, we use VGGFace2 \cite{cao2018vggface2} FRS to compare between the processing sample and each of the accepted samples in the dataset with a cosine-distance threshold of 0.45.
Finally, based on manual classification, we add pseudo-binary labels to the sampled base images as their gender and roughly classify and select 1175 male and 1175 female base samples to reduce the bias of the dataset. Then, the samples are divided into train, dev, and test sets for the convenience of training deep learning algorithms. 

\subsection{Algorithm of generating mated samples for SynMorph dataset}
To generate mated samples, given a base sample, we generate the mated sample of this subject by editing identity-irrelevant attributes, e.g., illumination, ageing effect, etc., based on pre-computed latent shifting directions \cite{colbois2021use} \cite{shen2020interfacegan}. Meanwhile, different editing strategies are applied to simulate the data used in different application scenarios (S-MAD or D-MAD). 
Face editing is, similar to the editing during the neutralization process when generating base samples, achieved by linearly interpolating the latent vector used to generate the face image on a specific direction and scale factor. The directions are pre-computed decision boundaries of semantic face attributes in the latent space and the scale factor is a scalar controlling the scale of editing (for example, a larger scale factor for age progression will add stronger ageing effects on the edited face image). For the S-MAD case, to keep the face image quality acceptable for passports, we edited the combination of illumination and ageing effect in a minor scale (noted as IFGS - InterFaceGAN for S-MAD). Given a normalized $\mathcal{W}$-latent vector $w^B$ from the base samples and unit normal vectors from the decision hyperplane of illumination flashed from left to the right $\hat{n}_I$, the ageing effect from younger or older than 30 years old $\hat{n}_A$ following: $w^{IFGS} = w^B + \alpha_{I} \cdot \hat{n}_I + \alpha_{A} \cdot \hat{n}_A$ with different combinations of scale factors $\alpha_{I}$ and $\alpha_{A}$.
For the D-MAD case, to simulate the probe images at the gate, editing to simulate the wilder condition is required. In the setting named IFGD, we edit more attributes, including pose, expression, illumination, and ageing effect, with larger scale factors $\beta$ as: $w^{IFGD} = w^B + \beta_P \cdot \hat{n}_P + \beta_{NS} \cdot \hat{n}_{NS} + \beta_{I} \cdot \hat{n}_I + \beta_{A} \cdot \hat{n}_A$. In the setting of FRPCA, we apply the random editing method proposed by Grimmer et al. \cite{grimmer2021generation} using PCA and control of VGGFace2 \cite{cao2018vggface2} FRS. 55 principle components in $\mathcal{W}$ latent space are computed as $\hat{n}_{PCA}^i$. For all of the generated mated samples, VGGFace2 FRS is applied to ensure the identity preservation between a base sample and generated mated samples.

\subsection{Generation of Morphs}
For the generation of morphs, we select one GAN-based morphing algorithm, MIPGAN-II \cite{zhang2021mipgan} and one landmark-based morphing algorithm, LMA-UBO \cite{UBO_Morphing_Tool}. To select the pairs of images for generating morphs, VGGFace2 \cite{cao2018vggface2} FRS model is used to compute the similarity score of each base image and other base images with the same gender and set (training, testing and validation). For the training set, 50 pairs with the top 50 highest similarity scores are selected. For the Dev and Test set, a full combination of pairs is selected due to the number of subjects. In this way, around 100000 morphs are generated for each morphing algorithm.

\section{Experiment and results}
\label{sec:experiment}
The objective of our experiment design is to evaluate the performance of the proposed SynMorph method for generating a synthetic morph dataset. By using the proposed method, we first generate a synthetic face morphing dataset with a large number of samples. Then, we evaluate the dataset from different perspectives and benchmark it with a non-synthetic face morphing dataset for further studies. As a face-morphing dataset, the two main performance factors are face image quality and attack ability towards FRS. Data with low face image quality might not be acceptable for the face recognition system and morphing attacks that are not able to threaten FRS are not effective and representative of attacks \cite{ISO-IEC-39794-5-G3-FaceImage-191015}. Also, it would be essential to compare with non-synthetic data and study consistency or gap between their performances. Finally, one of the intentions of developing the synthetic face morphing dataset is to use it as a large-scale and privacy-friendly dataset for developing and benchmarking morphing attack detection systems. Hence, we will select several S-MAD and D-MAD algorithms and benchmark them with different evaluation protocols using proposed synthetic and non-synthetic face morphing datasets.
\subsection{Dataset}
\begin{figure*}
    \centering
    \includegraphics[width=0.9\linewidth]{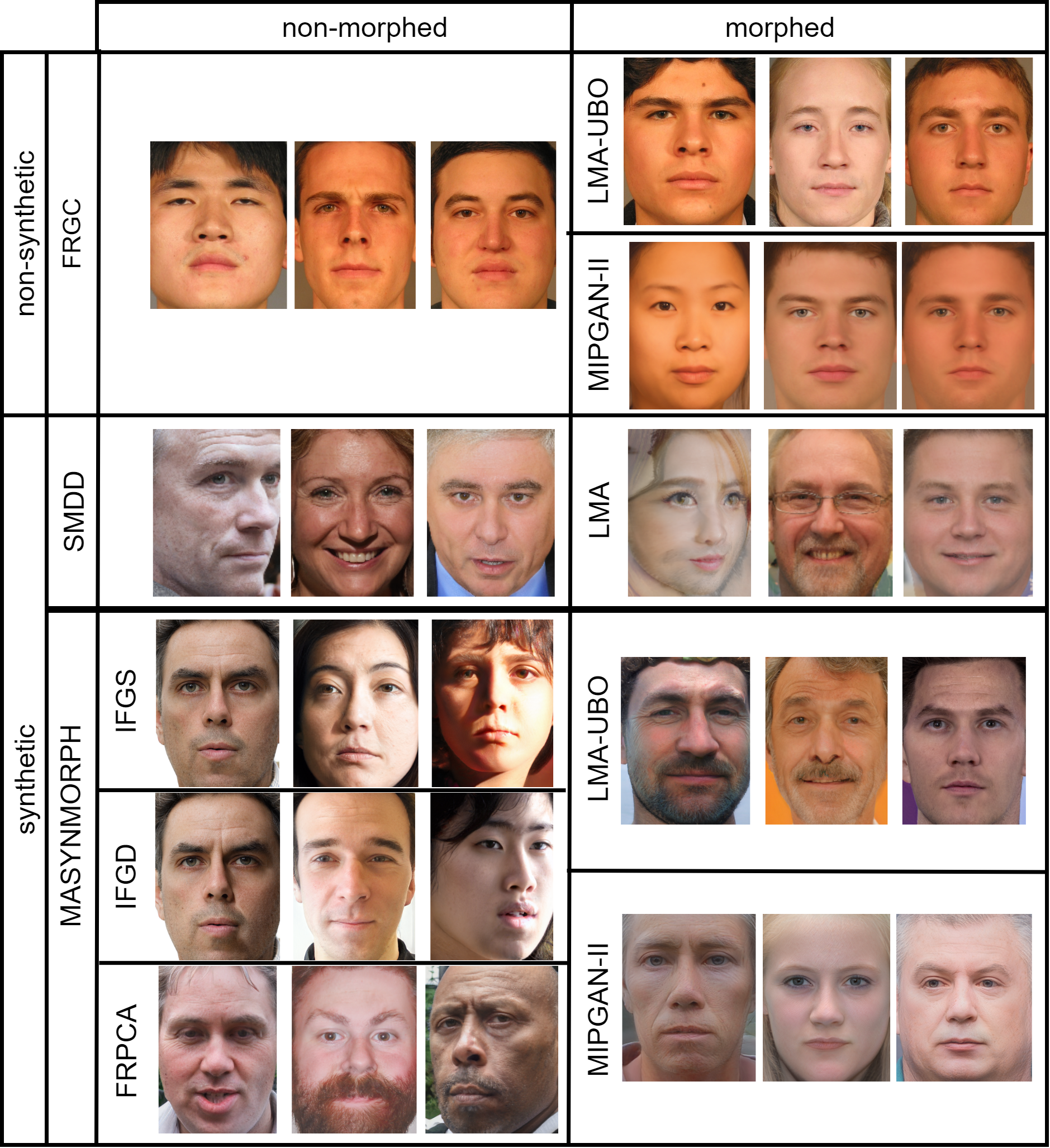}
    \caption{Overview of the generation of SynMorph dataset. Each triplet of images is selected based on SER-FIQ quality score: left-lowest, middle-median, right-highest. In D-MAD cases, IFGS images will be used as non-synthetic enrollment images, IFGD or FRPCA will be used as probe images with wilder capturing conditions.}
    \label{fig:examples}
\end{figure*}

As described in Section \ref{sec:proposed}, We first randomly generate base samples and manually select 1175 male and 1175 female as the 2350 different identities with binary pseudo gender in our dataset. Then, similar to constructing the non-synthetic face morphing dataset, we first split the base samples into training (1000), development (75), and testing (100) sets for the male and female groups, respectively. Then for each sub-group, we generate the mated samples and morphs. For each base sample, we generate a fixed number of mated samples for the three types of mated sample generation methods named IFGS (63), IFGD (90), and FRPCA (55). For the IFGS and IFGD methods, the generated mated samples are further filtered based on FRS to exclude the images without identity preservation, while the FRPCA algorithm itself manages the scale of editing using FRS control so there's no further filtering process. Finally, for the morph generating, we generate 50 morphs for each subject without duplications and symmetric pairs (Subject A is morphed with B, and again, subject B is morphed with A). The morph pairs are based on base samples from each group and hence without crossing of genders. As the development set and test set have 75 and 100 base samples, we use a full combination of the subject pairs to generate the morphs in order to keep a balanced number of morphed and non-morphed data. In the end, our SynMorph dataset 141k (IFGS), 210k (IFGD), and 129k (FRPCA) non-morphed samples, and 115k (each of MIPGAN \cite{zhang2021mipgan} and LMA \cite{UBO_Morphing_Tool}) morphed samples. The base images are generated by StyleGAN2 model \cite{karras2020analyzing} trained on the FFHQ \cite{karras2019style} dataset with an image resolution of 1024 by 1024.
To compare the quality between existing synthetic morph datasets, we selected the SMDD \cite{damer2022privacy} dataset as a baseline and also benchmark with a representative high-quality non-synthetic morph dataset based on FRGC V2 dataset \cite{zhang2021mipgan}. SMDD dataset (training part) contains 25k non-morphed images and 15k morphed images generated by landmark-based algorithm \cite{OpenCV_FaceMorph:2017}. As for the representative non-synthetic dataset, we select a high-quality and ICAO-compliant \cite{ICAO-9303-p1-2015} dataset \cite{zhang2021mipgan} based on FRGC v2 \cite{FRGC_DB} dataset. It includes 140 data subjects (47 female and 93 male) and each data subject has additional 7 to 21 mated samples, making the whole dataset 1270 non-morphed samples. For each of MIPGAN-II \cite{zhang2021mipgan} and LMA-UBO \cite{UBO_Morphing_Tool} morphing algorithm, around 2.5k morphs are generated.
Example images for each dataset are shown in Figure \ref{fig:examples}. Each triplet of images is selected based on SER-FIQ quality score: left-lowest, middle-median, right-highest.

\subsection{Face Image Quality Assessment}
As face morphing attack aims at attacking face recognition systems, it is essential to evaluate its biometric sample quality. Meanwhile, inspired by \cite{zhang2021applicability}, we measure the synthetic dataset's applicability by applying Face Image Quality Assessment (FIQA). 
Face Image Quality Assessment (FIQA) estimates the recognition performance of biometric systems. In this work, we selected FaceQnet v1\cite{hernandez2019faceqnet} and SER-FIQ\cite{terhorst2020ser} algorithms to extract the quality scores. FaceQnet v1 is an end-to-end deep learning model that is trained by labelling the FRS comparison score between to-be-estimated samples and high-quality samples as ground-truth scores. SER-FIQ is an unsupervised and FRS-dependent approach that estimates the quality score by applying dropout on a specific face recognition network to obtain its subnetworks and then measuring the stability of embeddings extracted by different sub-networks. Hence, it covers both supervised and unsupervised FIQA methods.

As a comparison, we select a representative non-synthetic face morphing dataset generated by FRGC v2 \cite{FRGC_DB} database using the same morphing algorithms as our synthetic dataset and with different pre-processing processes \cite{zhang2021mipgan}. 

More specifically, for the evaluation methodology, face quality scores of different types of data are extracted, and the score distortions will be qualitatively visualized in Kernel Density Estimate (KDE) plots and quantitatively measured by Kullback–Leibler divergences. 

\begin{figure}
    \centering
    \includegraphics[width=0.9\linewidth]{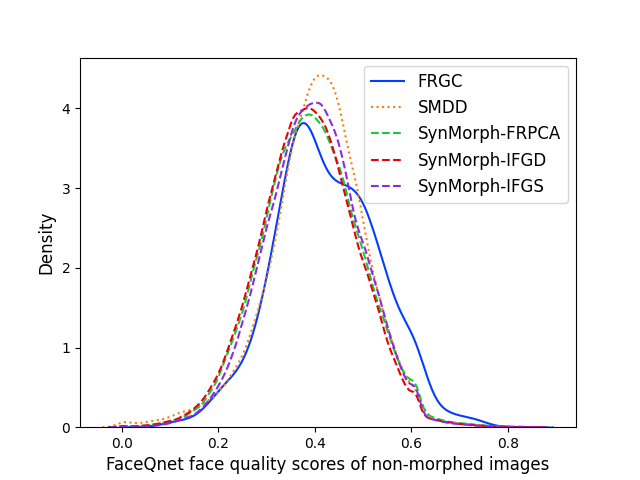}
    \caption{Distribution of FaceQnet face image quality scores of non-morphed images.}
    \label{fig:FaceQnet_bonafide}
\end{figure}

\begin{figure}
    \centering
    \includegraphics[width=0.9\linewidth]{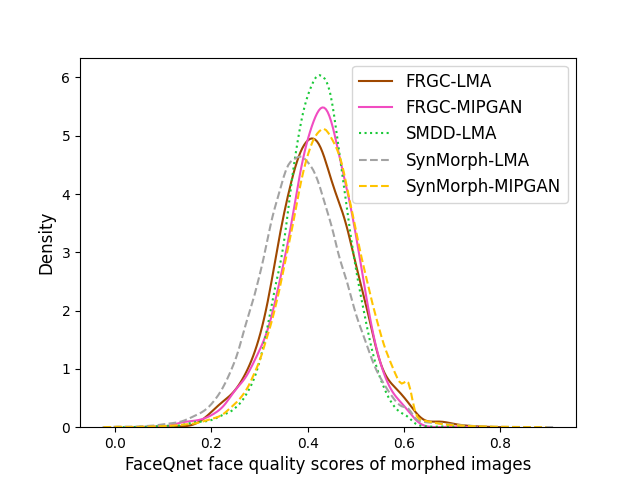}
    \caption{Distribution of FaceQnet image quality scores of morphed images.}
    \label{fig:FaceQnet_morphed}
\end{figure}

\begin{figure}

    \centering
    \includegraphics[width=0.9\linewidth]{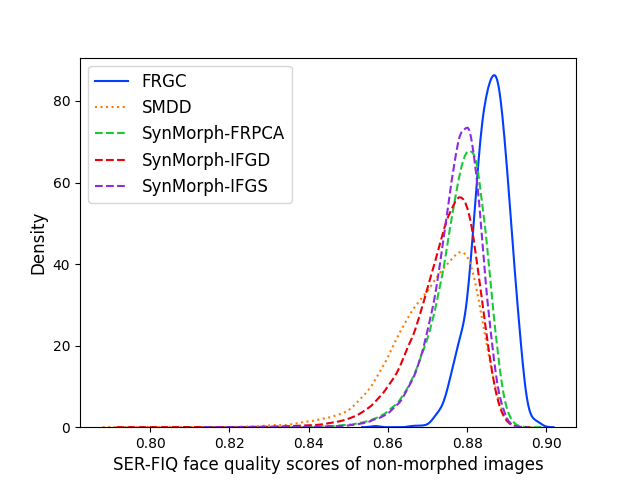}
    \caption{Distribution of SER-FIQ face image quality scores of non-morphed images.}
    \label{fig:SERFIQ_bonafide}
\end{figure}

\begin{figure}
    \centering
                                                           \includegraphics[width=0.9\linewidth]{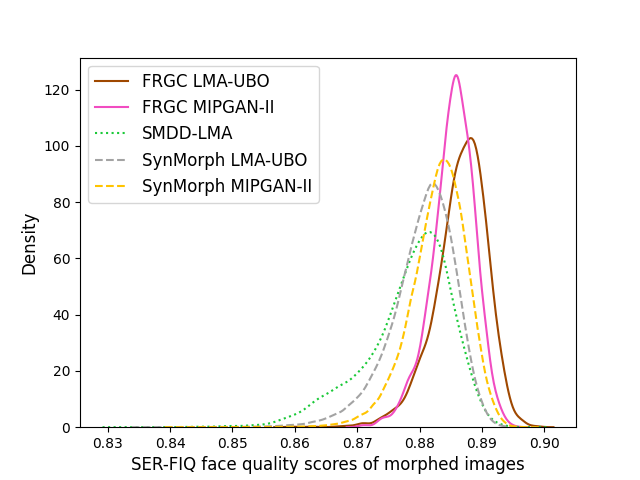}
    \caption{Distribution of SER-FIQ face image quality scores of morphed images.}
    \label{fig:SERFIQ_morphed}
\end{figure}

\begin{table*}[]
\centering
\begin{tabular}{|c|cc|cc}
\cline{1-3}
\multirow{2}{*}{FIQA} & \multicolumn{2}{c|}{Approaximate   Distritbution}    &  &  \\ \cline{2-3}
                      & \multicolumn{1}{c|}{SMDD}           & Proposed-IFGS       &  &  \\ \cline{1-3}
FaceQnet              & \multicolumn{1}{c|}{\textbf{0.087}} & 0.094          &  &  \\ \cline{1-3}
SER-FIQ                & \multicolumn{1}{c|}{2.799}          & \textbf{2.256} &  &  \\ \cline{1-3}
\end{tabular}
\caption{KL-D between the non-synthetic FRGC dataset and synthetic datasets: non-morphed images.}
\label{tab:FIQA_nonmorphed}
\end{table*}

\begin{table*}[]
\centering
\begin{tabular}{|c|ccc|c}
\cline{1-4}
\multicolumn{1}{|c|}{\multirow{2}{*}{FIQA}} & \multicolumn{3}{c|}{Approximate   Distribution}                                                    &  \\ \cline{2-4}
\multicolumn{1}{|c|}{}                      & \multicolumn{1}{c|}{SMDD-LMA} & \multicolumn{1}{c|}{Proposed-LMA} & Proposed-MIPGAN &  \\ \cline{1-4}
FaceQnet                                    & \multicolumn{1}{c|}{0.076}    & \multicolumn{1}{c|}{\textbf{0.075}}      & \textbf{0.038}            &  \\ \cline{1-4}
SER-FIQ                                      & \multicolumn{1}{c|}{1.657}    & \multicolumn{1}{c|}{\textbf{1.436}}      & \textbf{0.239}            &  \\ \cline{1-4}
\end{tabular}
\caption{KL-D between the non-synthetic FRGC dataset and synthetic datasets: morphed images.}
\label{tab:FIQA_morphed}
\end{table*}

Figure \ref{fig:FaceQnet_bonafide} and Figure \ref{fig:FaceQnet_morphed} illustrate the distributions of FaceQnet v1 scores for morphed and bona fide samples, respectively. It is shown that synthetic samples from both SMDD, the proposed SynMorph dataset, and the representative non-synthetic dataset have close distributions. Similar results can be also observed in Table \ref{tab:FIQA_morphed}.

However, for the results of SER-FIQ assessment, bona fide non-morphed images have shown, on average, the highest quality scores as shown in Figure \ref{fig:SERFIQ_bonafide} while the IFGS mated samples from the proposed dataset have shown higher quality compared to the baseline SMDD sample. For IFGD and FRPCA mated samples from the proposed SynMorph dataset, as they are aiming to represent the probe images in a D-MAD scenario with more variant attributes, the quality is lower than IFGS mated samples, as expected. For morphed images, the proposed method also shows a better quality than the SMDD samples. It also shows smaller KL-D in Table \ref{tab:FIQA_morphed} compared to the non-synthetic morph images

\subsection{Vulnerability Analysis}
\begin{figure*}[ht!]
    \centering
    \includegraphics[width=0.9\linewidth]{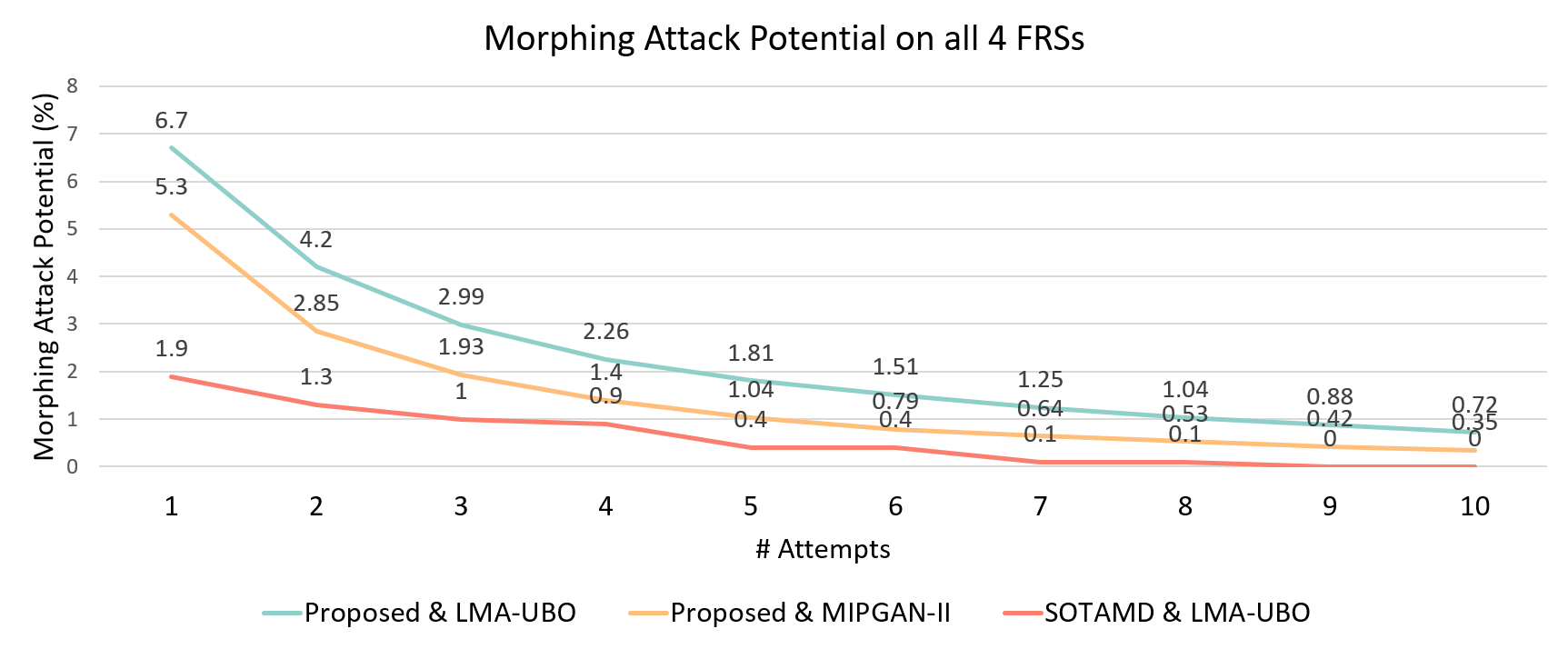}
    \caption{Visualisation on Morphing Attack Potential.}
\label{fig:MAP}
\end{figure*}

Vulnerability analysis on FRS against our SynMorph dataset, Morphing Attack Potential (MAP) \cite{ferrara2022morphing} is applied to measure the possibility of a successful morphing attack on multiple FRS with multiple mated samples. The metric is being standardized in ISO/IEC 20059 \cite{ISO_20059}. More specifically, 4 FRS implemented in Deepface library \cite{serengil2020lightface} are included for evaluation: ArcFace \cite{Deng_2019_arcface}, Dlib \cite{dlib}, Facenet \cite{schroff2015facenet}, and VGGFace \cite{parkhi2015deep}. The results of our SynMorph dataset are benchmarked with SMDD dataset. For the representative non-synthetic dataset, because the FRGC morphing dataset has a limited number of mated samples, we benchmark with the SOTAMD dataset \cite{raja2020morphing} from the original MAP paper (TABLE IX in \cite{ferrara2022morphing}). As shown in Figure \ref{fig:MAP}, the proposed method shows a considerable MAP, indicating its effectiveness on threatening FRS. Furthermore, the landmark-based method shows and higher attack potential than the GAN-based method. Overall, the MAP of the proposed SynMorph dataset is higher than the SOTAMD non-synthetic dataset.  

\subsection{Morphing Attack Detection}

For morphing attack detection experiments, we designed 3 protocols:
\begin{itemize}
    \item Protocol I: training and testing sets have the same type of data
    \item Protocol II: training and testing sets have different types of data
    \item Protocol III: training set is mixed with synthetic and non-synthetic data, tested on synthetic and non-synthetic data separately
\end{itemize}

Protocol I evaluates the common scenario of using the same type of data to construct the training and testing set of MAD algorithm. For example, if the training set contains non-synthetic data, the test set also contains non-synthetic data. This simulates the applicability of using synthetic data for benchmarking between MAD algorithms.
Further, it should be investigated whether the model trained on synthetic data can be generalized to the detection of non-synthetic data. Protocol II evaluates the MAD performance of cross-testing between data types. If the model is trained with synthetic data, the MAD results on non-synthetic data will be reported, and vice versa when the model is trained on non-synthetic data. Finally, in Protocol III we evaluate the impact of training MAD with both synthetic and non-synthetic data together. The testing results will be separately reported with only synthetic and non-synthetic data. In all these three protocols, cross-testing of morphing types is considered. 

As for selected algorithms to be trained, we select 2 S-MAD algorithms and 2 D-MAD algorithms:
\begin{itemize}
    \item MorphHRNet: A end-to-end S-MAD algorithm based on HRNet \cite{wang2020deep} model, submitted to SYN-IJCB-2022 \cite{huber2022syn}.
    \item Xception: A end-to-end S-MAD algorithm based on Xception model \cite{chollet2017xception}, submitted in SYN-IJCB-2022 \cite{huber2022syn}.
    \item Differential Deep Face Representations (DDFR) \cite{scherhag2020deep}: A D-MAD approach based on the difference of features extracted using ArcFace FRS network. Classification is done by a linear SVM classifier.
    \item Landmark-based Face De-morphing (LMFD) \cite{ferrara2017face}: The inverse process of landmark-based morphing between suspicious enrollment images and probe images, MAD is conducted by verification of de-morphed images and probe images.
\end{itemize}
For MorphHRNet, Xception, and DDFR algorithms, Detection Error Tradeoff curves of Morphing Attack Classification Error Rate (MACER) and Bona fide Presentation Classification Error Rate(BPCER) will be plotted to visualize the results \cite{ISO_20059}. As the name indicates, MACER measures the possibility of morphing attacks being misclassified as bona fide presentations, and BPCER measures the possibility of bona fide presentations being classified as morphing attacks. 
For the LMFD algorithm, the original algorithm is based on a two-step classification and outputs a binary classification result instead of a score: first verify between the suspicious image and the probe image, and then verify between the de-morphed image and the probe image. To keep consistency between other benchmarking by ploting the DET curve, we only use the second step: verify between the de-morphed image and the probe image. Empirically, a 0.5 factor is used for de-morphing and the ArcFace FRS \cite{Deng_2019_arcface} model is applied to extract the face embeddings. Furthermore, as it is based on landmark-based de-morphing and FRS comparison, there's no training set for the benchmarking and hence no different evaluation protocols crossing between the training set and testing sets when evaluating LMFD. Due to the efficiency of the de-morphing algorithm, during the evaluation of synthetic dataset using LMFD, we only randomly selected one IFGD sample and one FRPCA sample for each subject.

\subsection{Evaluation on S-MAD Algorithms}
\begin{figure*}[htp]
     \centering
     \begin{subfigure}[b]{0.24\textwidth}
         \centering
         \includegraphics[width=\textwidth]{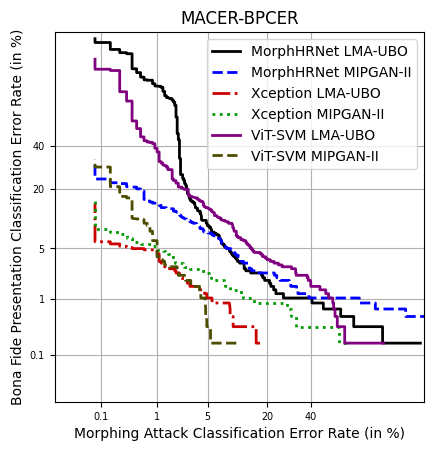}
         \caption{Trained on Non-Syn. data with LMA-UBO morphs}
         \label{fig:SMAD_1_1}
     \end{subfigure}
     \hfill
     \begin{subfigure}[b]{0.24\textwidth}
         \centering
         \includegraphics[width=\textwidth]{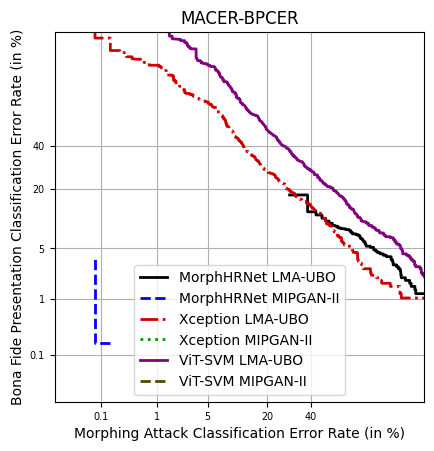}
         \caption{Trained on Non-Syn. data with MIPGAN-II morphs}
         \label{fig:SMAD_1_2}
     \end{subfigure}
     \hfill
     \begin{subfigure}[b]{0.24\textwidth}
         \centering
         \includegraphics[width=\textwidth]{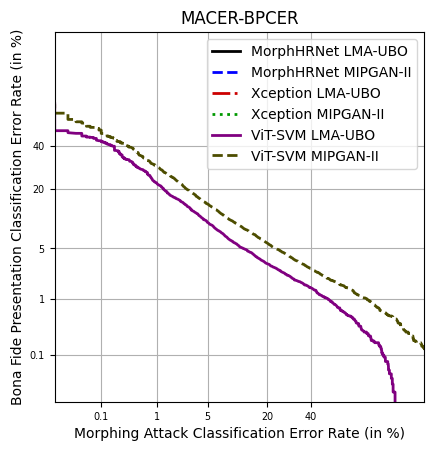}
         \caption{Trained on Syn. data with LMA-UBO morphs}
         \label{fig:SMAD_1_3}
     \end{subfigure}
     \hfill
     \begin{subfigure}[b]{0.24\textwidth}
         \centering
         \includegraphics[width=\textwidth]{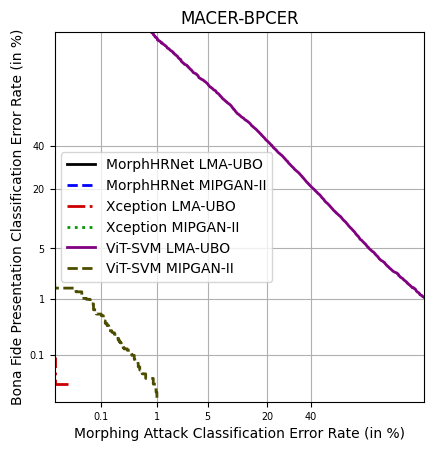}
         \caption{Trained on Syn. data with MIPGAN-II morphs}
         \label{fig:SMAD_1_4}
     \end{subfigure}
        \caption{S-MAD results of MorphHRNET and Xception: Training and testing sets have the same type of data (synthetic or non-synthetic). Trained models are tested on morphs generated with different morphing algorithms.}
        \label{fig:SMAD_1}
\end{figure*}

Figure \ref{fig:SMAD_1} shows the evaluation results of Protocol I on S-MAD algorithms. As noted in Figure \ref{fig:SMAD_1_1} and Figure \ref{fig:SMAD_1_2}, when trained with non-synthetic LMA-UBO data, the Xception method achieves a lower detection error rate than the MorphHRNet Algorithm. It can also be noticed that training with LMA-UBO morphs achieves higher generalizability than training with MIPGAN-based morphs, while training and testing with MIPGAN-II morphs are easier than the LMA-UBO morphs. Figure \ref{fig:SMAD_1_3} and \ref{fig:SMAD_1_4} shows that when using large numbers of synthetic data for training and testing, both algorithms have shown very low classification error rate even for detecting the unknown type of morphing attacks. 

\begin{figure*}[htp]
     \centering
     \begin{subfigure}[b]{0.24\textwidth}
         \centering
         \includegraphics[width=\textwidth]{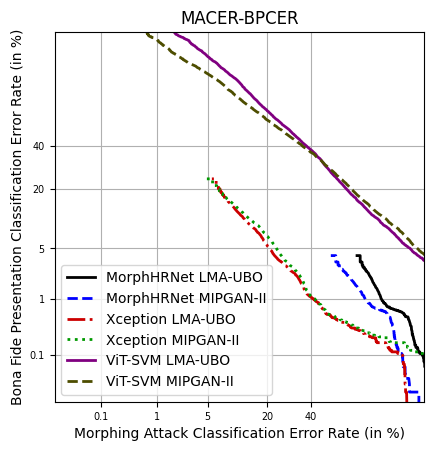}
         \caption{Trained on Non-Syn. data with LMA-UBO morphs}
         \label{fig:SMAD_2_1}
     \end{subfigure}
     \hfill
     \begin{subfigure}[b]{0.24\textwidth}
         \centering
         \includegraphics[width=\textwidth]{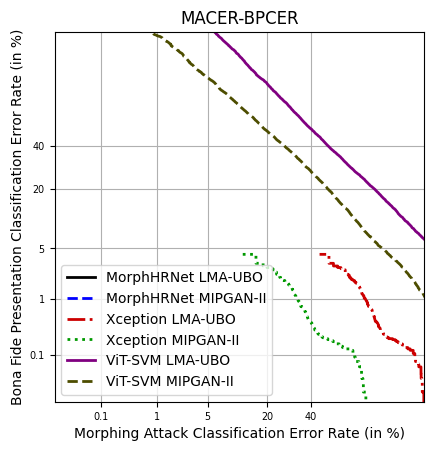}
         \caption{Trained on Non-Syn. data with MIPGAN-II morphs}
         \label{fig:SMAD_2_2}
     \end{subfigure}
     \hfill
     \begin{subfigure}[b]{0.24\textwidth}
         \centering
         \includegraphics[width=\textwidth]{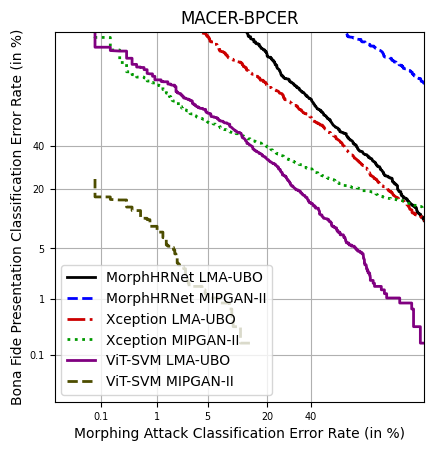}
         \caption{Trained on Syn. data with LMA-UBO morphs}
         \label{fig:SMAD_2_3}
     \end{subfigure}
     \hfill
     \begin{subfigure}[b]{0.24\textwidth}
         \centering
         \includegraphics[width=\textwidth]{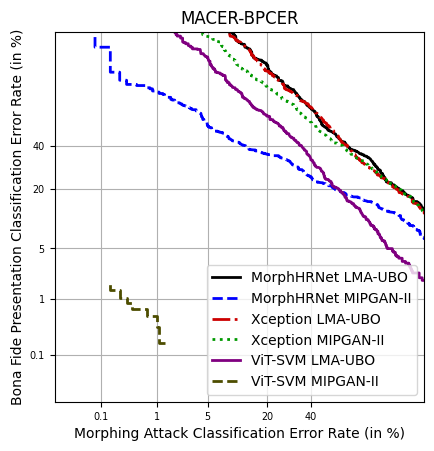}
         \caption{Trained on Syn. data with MIPGAN-II morphs}
         \label{fig:SMAD_2_4}
     \end{subfigure}
     
        \caption{S-MAD results of MorphHRNET and Xception: Training and testing sets have different types of data (synthetic or non-synthetic). Trained models are tested on morphs generated with different morphing algorithms.}
        \label{fig:SMAD_2}
\end{figure*}

As shown in Figure \ref{fig:SMAD_2_1} and Figure \ref{fig:SMAD_2_2}, when training on non-synthetic data and testing on synthetic data, the Xception algorithm shows a more robust performance than MorphHRNet. Both algorithms show a quite high error rate when training on synthetic data and testing on non-synthetic data.

\begin{figure*}[htp]
     \centering
     \begin{subfigure}[b]{0.24\textwidth}
         \centering
         \includegraphics[width=\textwidth]{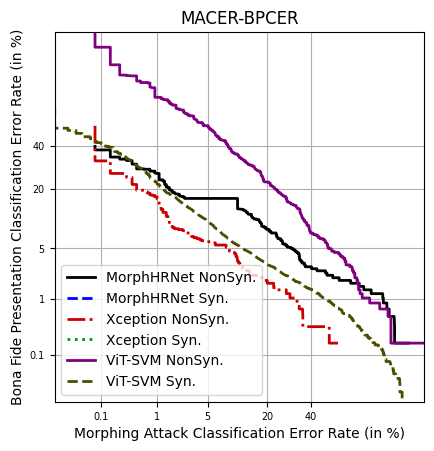}
         \caption{Train: LMA-UBO\\Test:LMA-UBO}
         \label{fig:SMAD_3_1}
     \end{subfigure}
     \hfill
     \begin{subfigure}[b]{0.24\textwidth}
         \centering
         \includegraphics[width=\textwidth]{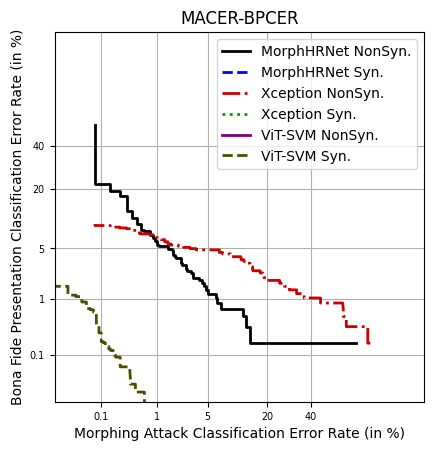}
         \caption{Train: MIPGAN-II\\Test:MIPGAN-II}
         \label{fig:SMAD_3_2}
     \end{subfigure}
     \hfill
     \begin{subfigure}[b]{0.24\textwidth}
         \centering
         \includegraphics[width=\textwidth]{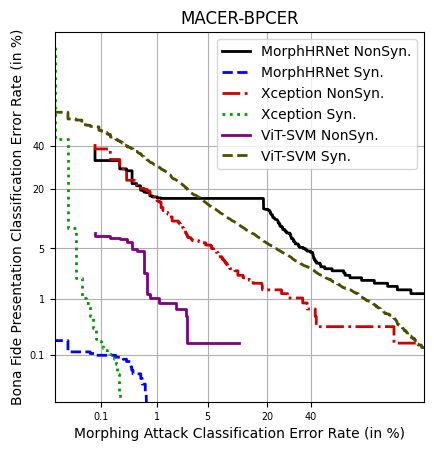}
         \caption{Train: LMA-UBO\\Test:MIPGAN-II}
         \label{fig:SMAD_3_3}
     \end{subfigure}
     \hfill
     \begin{subfigure}[b]{0.24\textwidth}
         \centering
         \includegraphics[width=\textwidth]{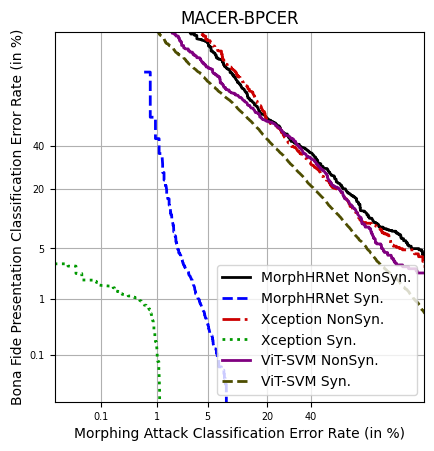}
         \caption{Train: MIPGAN-II\\Test:LMA-UBO}
         \label{fig:SMAD_3_4}
     \end{subfigure}

        \caption{S-MAD results of MorphHRNET and Xception: trained with synthetic and non-synthetic together, and tested on different types of data generated by different morphing algorithms.}
        \label{fig:SMAD_3}
\end{figure*}

In Figure \ref{fig:SMAD_3}, we report the Protocol III results when MAD algorithms are trained with together synthetic and non-synthetic data and have different train-test settings. An overall lower detection error rate on synthetic data can be noticed due to the larger size of synthetic data compared to non-synthetic data in the training set. Comparing the benchmarked algorithms to detect non-synthetic morphing algorithms, Figure \ref{fig:SMAD_3_1} shows that Xception algorithm has a higher accuracy when the model is trained by LMA-UBO morphs, while Figure \ref{fig:SMAD_3_2} indicates that the MorphHRNet performs better BPCER at low MACER. Regarding the different sessions in Protocol III, cross-testing between morphing algorithms, in general, increases the error rate. Comparing training with the same type of data and training with combined types of data, for example, curve 'Xception NonSyn.' in Figure \ref{fig:SMAD_3_1} with the curve labelled as 'Xception LMA-UBO' in Figure \ref{fig:SMAD_1_1} and \ref{fig:SMAD_1_3}, the classification error rate is in the same level but slightly higher. While comparing to having different types of training data and different types of morphs in Figure \ref{fig:SMAD_2}, introducing both types of data during training will reduce the classification error rate on non-synthetic data.


\subsection{Evaluation on D-MAD Algorithms}
\begin{figure*}[htp]
     \centering
     \begin{subfigure}[b]{0.3\textwidth}
         \centering
         \includegraphics[width=\textwidth]{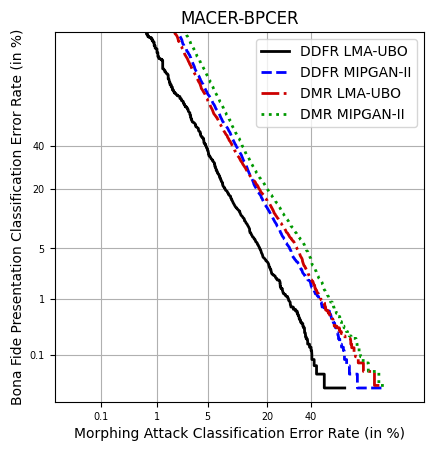}
         \caption{Trained on Non-Syn. data with LMA-UBO morphs}
         \label{fig:DMAD_1_1}
     \end{subfigure}
     \begin{subfigure}[b]{0.3\textwidth}
         \centering
         \includegraphics[width=\textwidth]{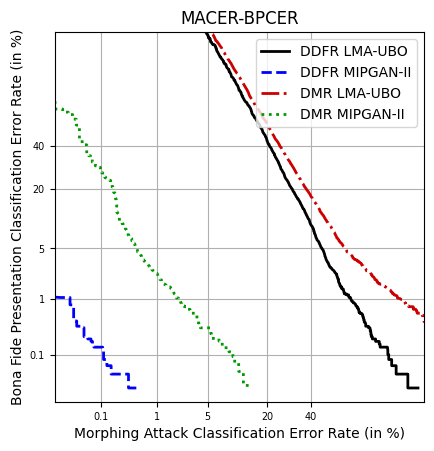}
         \caption{Trained on Non-Syn. data with MIPGAN-II morphs}
         \label{fig:DMAD_1_2}
     \end{subfigure}
     \begin{subfigure}[b]{0.3\textwidth}
         \centering
         \includegraphics[width=\textwidth]{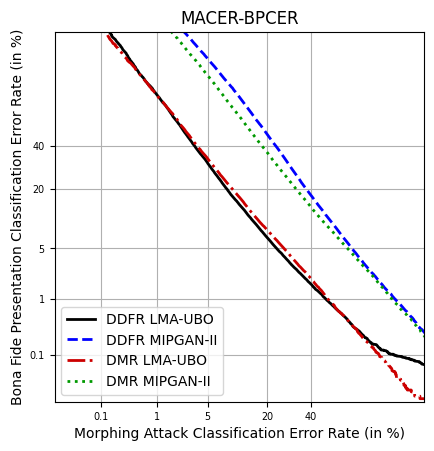}
         \caption{Trained on Syn. IFGD data with LMA-UBO morphs}
         \label{fig:DMAD_1_3}
     \end{subfigure}
     \begin{subfigure}[b]{0.3\textwidth}
         \centering
         \includegraphics[width=\textwidth]{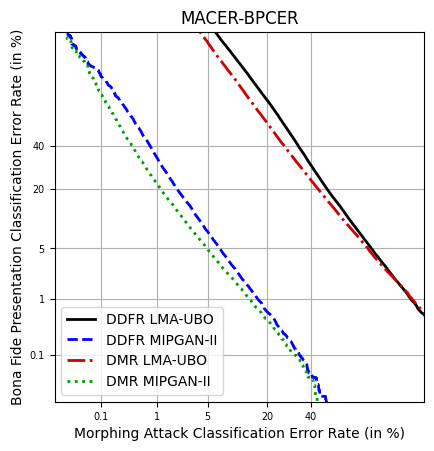}
         \caption{Trained on Syn. IFGD data with MIPGAN-II morphs}
         \label{fig:DMAD_1_4}
     \end{subfigure}
     \begin{subfigure}[b]{0.3\textwidth}
         \centering
         \includegraphics[width=\textwidth]{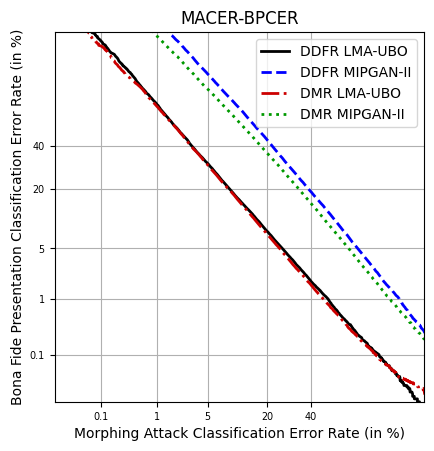}
         \caption{Trained on Syn. FRPCA data with LMA-UBO morphs}
         \label{fig:DMAD_1_5}
     \end{subfigure}
     \begin{subfigure}[b]{0.3\textwidth}
         \centering
         \includegraphics[width=\textwidth]{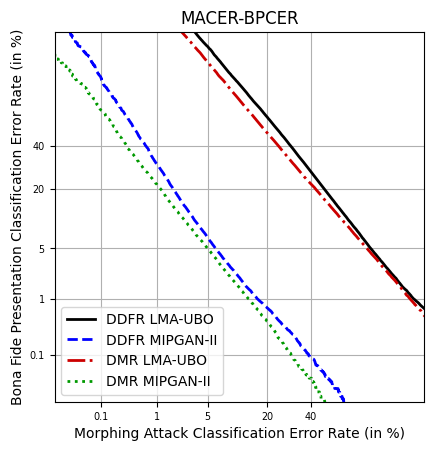}
         \caption{Trained on Syn. FRPCA data with MIPGAN-II morphs}
         \label{fig:DMAD_1_6}
     \end{subfigure}
        \caption{D-MAD results of DDFR algorithm: Training and testing sets have the same type of data (synthetic or non-synthetic). Trained models are tested on morphs generated with different morphing algorithms.}
        \label{fig:DMAD_1}
\end{figure*}

\begin{figure*}[htp]
     \centering
     \begin{subfigure}[b]{0.3\textwidth}
         \centering
         \includegraphics[width=\textwidth]{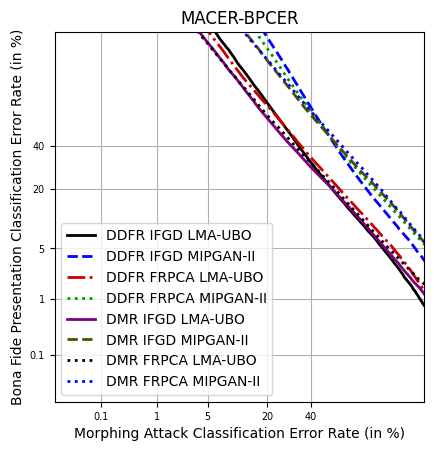}
         \caption{Trained on Non-Syn. data with LMA-UBO morphs}
         \label{fig:DMAD_2_1}
     \end{subfigure}
     \begin{subfigure}[b]{0.3\textwidth}
         \centering
         \includegraphics[width=\textwidth]{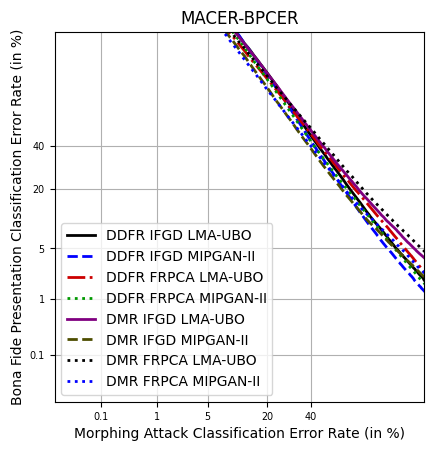}
         \caption{Trained on Non-Syn. data with MIPGAN-II morphs}
         \label{fig:DMAD_2_2}
     \end{subfigure}
    \begin{subfigure}[b]{0.3\textwidth}
         \centering
         \includegraphics[width=\textwidth]{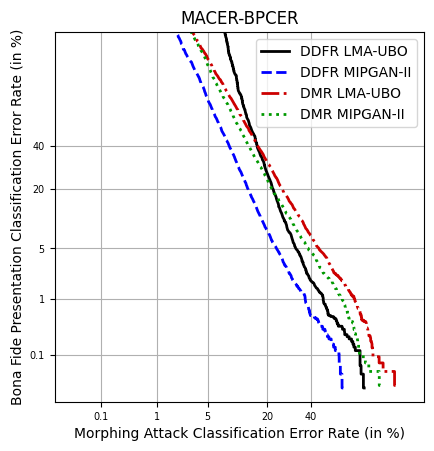}
         \caption{Trained on Syn. IFGD data with LMA-UBO morphs}
         \label{fig:DMAD_2_3}
     \end{subfigure}
     \begin{subfigure}[b]{0.3\textwidth}
         \centering
         \includegraphics[width=\textwidth]{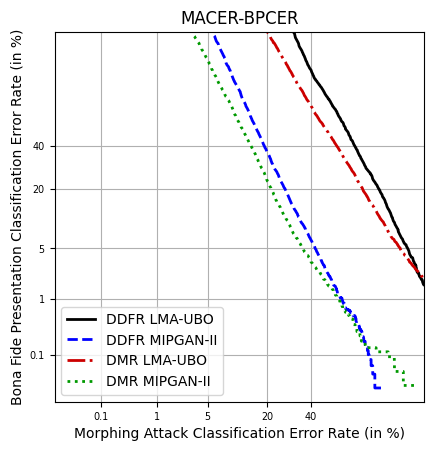}
         \caption{Trained on Syn. IFGD data with MIPGAN-II morphs}
         \label{fig:DMAD_2_4}
     \end{subfigure}
     \begin{subfigure}[b]{0.3\textwidth}
         \centering
         \includegraphics[width=\textwidth]{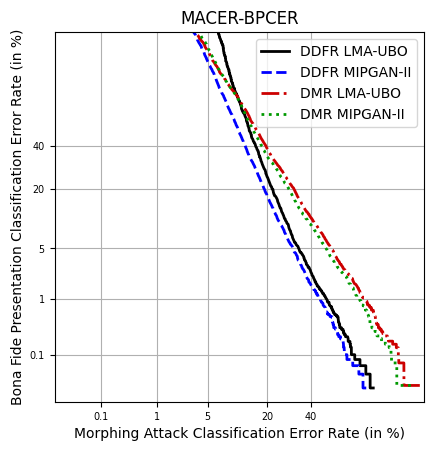}
         \caption{Trained on Syn. FRPCA data with LMA-UBO morphs}
         \label{fig:DMAD_2_5}
     \end{subfigure}
     \begin{subfigure}[b]{0.3\textwidth}
         \centering
         \includegraphics[width=\textwidth]{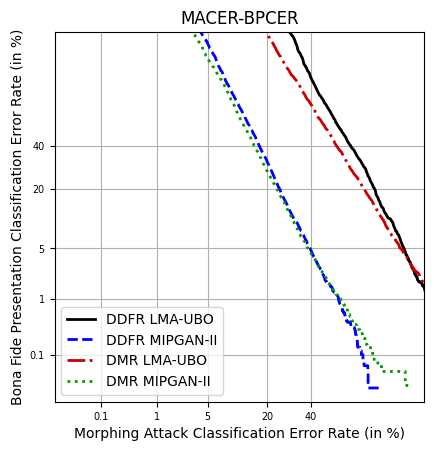}
         \caption{Trained on Syn. FRPCA data with MIPGAN-II morphs}
         \label{fig:DMAD_2_6}
     \end{subfigure}
        \caption{D-MAD results of DDFR algorithm: Training and testing sets have different types of data (synthetic or non-synthetic). Trained models are tested on morphs generated with different morphing algorithms.}
        \label{fig:DMAD_2}
\end{figure*}

\begin{figure*}
     \centering
     \begin{subfigure}[b]{0.24\textwidth}
         \centering
         \includegraphics[width=\textwidth]{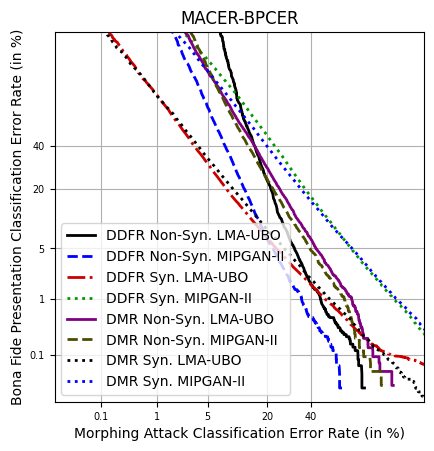}
         \caption{Trained on Non-Syn. \& Syn. IFGD data with LMA-UBO morphs}
         \label{fig:DMAD_3_1}
     \end{subfigure}
     \hfill
     \begin{subfigure}[b]{0.24\textwidth}
         \centering
         \includegraphics[width=\textwidth]{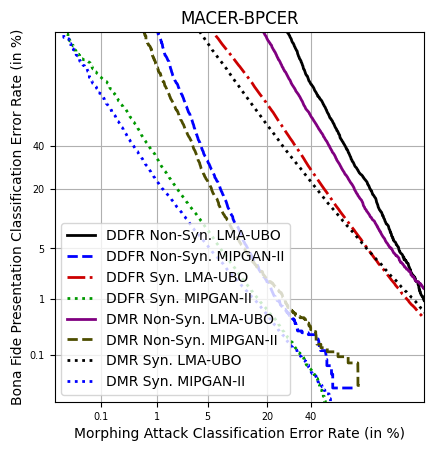}
         \caption{Trained on Non-Syn. \& Syn. IFGD data with MIPGAN-II morphs}
         \label{fig:DMAD_3_2}
     \end{subfigure}
     \hfill
     \begin{subfigure}[b]{0.24\textwidth}
         \centering
         \includegraphics[width=\textwidth]{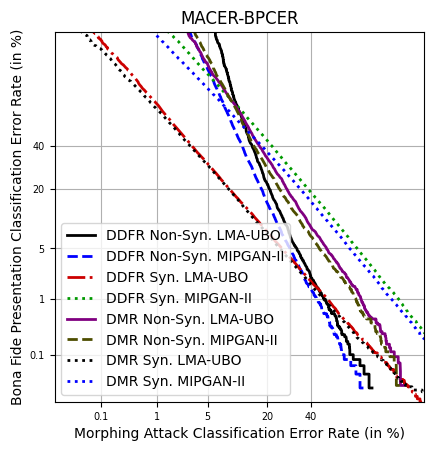}
         \caption{Trained on Non-Syn. \& Syn. FRPCA data with LMA-UBO morphs}
         \label{fig:DMAD_3_3}
     \end{subfigure}
     \hfill
     \begin{subfigure}[b]{0.24\textwidth}
         \centering
         \includegraphics[width=\textwidth]{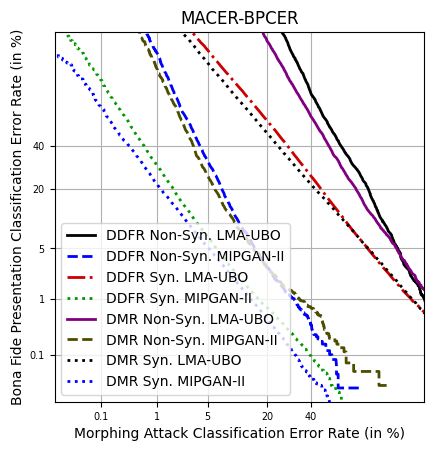}
         \caption{Trained on Non-Syn. \& Syn. FRPCA data with MIPGAN-II morphs}
         \label{fig:DMAD_3_4}
     \end{subfigure}
        \caption{D-MAD results of DDFR algorithm: trained with synthetic and non-synthetic together, and tested on different types of data generated by different morphing algorithms.}
        \label{fig:DMAD_3}
\end{figure*}

Differing from the previous evaluations of S-MAD algorithms, for D-MAD data we have two types of synthetic data: synthetic IFGD and synthetic FRPCA. Synthetic IFGS data will be used to simulate the enrollment non-morph data. Figure \ref{fig:DMAD_1} shows the D-MAD evaluation results of the DDFR algorithm for Protocol I. When training and testing on the same type of data, in all cases, the detection performance degradation on unknown attacks remains and is especially obvious in Figure \ref{fig:DMAD_1_2} where the model is generalizing with training data of non-synthetic data with MIPGAN-II morphs to LMA-UBO morphs. 
Evaluation results of training and testing with different types of data are illustrated in Figure \ref{fig:DMAD_2}. Similar to the observation for S-MAD results, cross-testing between different types of data, both for training on synthetic and testing on non-synthetic, and training on non-synthetic and testing on synthetic, shows a high classification error rate. It is also shown in Figure \ref{fig:DMAD_2_3} and \ref{fig:DMAD_2_5} that when the models are trained with synthetic LMA-UBO data and tested on non-synthetic data, results of inter-morphing-algorithm testing is similar and even lower than intra-morphing algorithm cases. In the comparison between Figure \ref{fig:DMAD_2_3}-\ref{fig:DMAD_2_4} and Figure \ref{fig:DMAD_2_5}-\ref{fig:DMAD_2_6}, training and testing the model with both MIPGAN-II based data even achieved lower detection accuracies.
The results of training with the mix of synthetic and non-synthetic data are shown in Figure \ref{fig:DMAD_3}. When the algorithm is trained with landmark-based morphs, it is shown in Figure \ref{fig:DMAD_3_1} that the classification error rate of testing on synthetic IFGD data with MIPGAN morphs is quite high. Other three curves when testing on non-synthetic data (with landmark-based or GAN-based morphs) have shown similar detection performance at low MACER. Figure \ref{fig:DMAD_3_2} shows the results of using MIPGAN-II morphs for training. In this case, testing on datasets with also MIPGAN-II morphs shows an overall lower classification error rate than datasets with LMA-UBO morphs. Comparing different types of testing data, detection accuracy on synthetic data is, in general, lower than results on non-synthetic data. Similar observations hold for using synthetic data with mated samples generated by the other algorithm (FRPCA) in  Figure \ref{fig:DMAD_3_3} and Figure \ref{fig:DMAD_3_4}.

\begin{figure}[htp]
    \centering
    \includegraphics[width=0.4\textwidth]{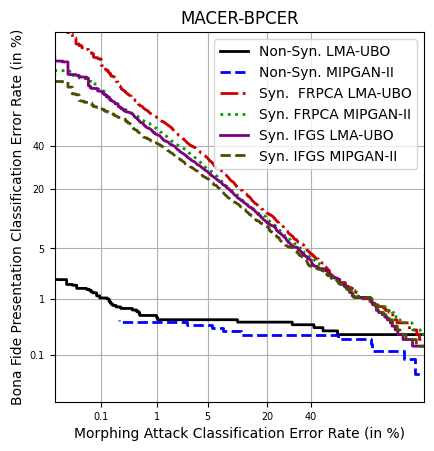}
    \caption{D-MAD results of LMFD algorithm.}
\label{fig:demorphing}
\end{figure}

Figure \ref{fig:demorphing} includes the benchmarking of the landmark-based face de-morphing algorithm (LMFD). It is shown that the MACER and BPCER of the synthetic data are overall higher than the non-synthetic data. Comparing results on using the same type of data but with morphs generated by different morphing algorithms, a consistent trend can be observed: detecting the MIPGAN-II morphs has a lower error rate than detecting LMA-UBO morphs. 

\section{Discussions and limitations}
\label{sec:discussion}
Our evaluation results on face image quality assessment show that the synthetic face morphing dataset also has a considerable face image quality, meaning that their quality is acceptable for a passport enrolment application and similar to non-synthetic data. Similar trends have also been indicated in other works for synthetic face data. For the SER-FIQ method, there's a gap between our method and the selected baseline synthetic dataset, SMDD dataset, and the non-synthetic face morphing dataset. This might be because the SMDD dataset is filtered based on FaceQnet quality scores during dataset generation. Hence the data show high-quality scores when again being evaluated by FaceQnet afterwards. In this case, our proposed method shows a higher face image quality than SMDD dataset and is also closer to the score distribution of non-synthetic data.

Regarding vulnerability analysis, the proposed SynMorph dataset shows a higher face morphing attack potential compared to the non-synthetic face morphing dataset, which shows the effectiveness of generated synthetic morphs. However, it should be noted that the vulnerability analysis is based on the comparisons between morphs and mated samples. For the synthetically generated samples, we used FRS-control for identity preservation and several editing techniques, while compared to the real application cases, it remains a challenge for the synthetic data to simulate the large variation on different mated face representations, especially for the probe images used for D-MAD with wilder capturing conditions. 

When benchmarking S-MAD algorithms, due to the small number of non-synthetic images for training, using MIPGAN-based morphs as training data usually makes it easy for the model to overfit on the determination between GAN-generated images and non-synthetic images instead of learning the traces of morphing, which makes it challenging to generalize on unseen LMA-based attacks. When training the S-MAD model with non-synthetic data with MIPGAN-based morphs, we used reconstructed non-morphed images with the same backbone StyleGAN2 generator as MIPGAN-II to mitigate the bias between non-morphed non-synthetic images and morphed non-synthetic images. However, the gap remains quite noticeable compared to models trained on the non-synthetic landmark-based dataset. This is also explained in Figure \ref{fig:SMAD_2_1}-\ref{fig:SMAD_2_2} where the MACER is very high. For results of Protocol II as shown in Figure \ref{fig:SMAD_2}, it is challenging for the algorithms trained only on non-synthetic data to directly generalize to synthetic data (or vice versa). On the other hand, for the synthetic data, as the non-morphed images of synthetic data are originally GAN-generated, the morphs generated by the landmark algorithm may also leave some GAN-based traces. Hence, when using the synthetic data for training, it is challenging to generalize on non-synthetic data as shown in Figure \ref{fig:SMAD_2_3}-\ref{fig:SMAD_2_4}. When training together with synthetic and non-synthetic data, the classification error rate reduces significantly compared to training with one type of data and testing one another, but also higher than training and testing with the same single type of data (intra-type evaluation).

For D-MAD cases, as the training pairs can be combinations of pairs with suspicious images and mated probe images, the training data are more sufficient and the reconstruction trick is not applied. It shows a larger gap when there is training and testing on the same type of data but different types of morphs, especially in Figure \ref{fig:DMAD_1_2} when the model is trained on non-synthetic MIPGAN-II data. The protocol II evaluation results also show that training on one type of data and testing on another is challenging. For the evaluation results of the identity-based LMFD algorithm in Figure \ref{fig:demorphing}, it is shown that the landmark-based face de-morphing method is also working for the synthetic dataset. As the de-morphing-based method is sensitive to the quality and condition of probe images captured in ABC-gates at border control, the non-synthetic data with high-quality probes achieved low MACER and BPCER. Comparing the two types of synthetic data generated by two different face editing algorithms, IFGD and FRPCA, the FRPCA-based method introduces more random variants and leads to a higher BPCER. However, the differences between the two types of synthetic data are not obvious in the benchmarking results of the DDFR algorithm. 

In general, we have shown that the SynMorph data can be used for benchmarking training and testing MAD algorithms. However, there remain differences between synthetic and non-synthetic data, which make it challenging for algorithms to generalize from one type of data to another.

Regarding the limitations of the SynMorph dataset, given the common scenario of face morphing attacks at automatic border control, usually, samples of different genders are not selected as morph pairs because the malicious attack will need to present as another gender than the document. In this case, we manually sorted the data into two bins of genders. These soft labels may also be done by gender classification, while the classification accuracy of implementations we tested was less satisfying. When generating base samples, the SynMorph method uses a loop with acceptance conditions based on checking the explicit quality measure and identity diversity. With the increasing number of accepted samples, the rejection rate also increases and makes the speed of generating base samples slower in the late stage.

For privacy concerns, we use randomly generated subjects and aim to make it privacy-friendly and convenient to researchers for benchmarking. Relevant research on privacy regulations regarding synthetic data to avoid privacy leakage remains an important topic to be studied. 

\section{Conclusion}
\label{sec:conclusion}

In this paper, we've proposed a new method for generating a synthetic face morphing dataset with high image quality and support for both S-MAD and D-MAD by generating the mated samples. Then, we use the proposed method to generate a large-scale synthetic morph dataset and evaluate its performance. Results show a higher face image quality compared to the baseline and considerably higher morphing attack potential to 4 FRS. Additionally, we studied the applicability of using our synthetic face morphing dataset for training S-MAD and D-MAD algorithms. Results show that the synthetic data can be used for training and evaluating MAD algorithms. Due to the large number of samples, generalizability between different types of MAs can be improved in some cases. However, it is also shown that crossing between bona fide and synthetic data remains challenging. Hence, it is suggested to carefully report when using synthetic data for evaluating MAD. It remains an open topic on how to effectively use synthetic face morphing datasets such as SynMorph to reduce the detection error rate of MAD algorithms on non-synthetic data.

\section{Acknowledgement}
This work was supported by the Image Manipulation Attack Resolving Solutions (iMARS) project which has received funding from the European Union’s Horizon 2020 Research and Innovation Program under Grant 883356.


%



\ifCLASSOPTIONcompsoc
  


\ifCLASSOPTIONcaptionsoff
  \newpage
\fi



%

\bibliographystyle{IEEEtran}
	\bibliography{references}




%








\end{document}